\documentclass[utf8]{frontiersSCNS}
\usepackage{graphics}
\usepackage{lineno}
\usepackage{amsmath}
\usepackage{amssymb}
\usepackage{multicol}
\usepackage{graphicx}
\usepackage{layouts}
\usepackage{wrapfig}
\usepackage{caption}
\usepackage{graphicx}
\usepackage{easyReview}
\usepackage{printlen}
\usepackage{bm}
\usepackage{siunitx}
\usepackage{colortbl}
\usepackage{hyperref}

\usepackage[onehalfspacing]{setspace}

\def\keyFont{\fontsize{8}{11}\helveticabold }
\def\firstAuthorLast{Bogdanovic {et~al.}}
\def\Authors{Miroslav Bogdanovic\,$^{1,*}$, Majid Khadiv\,$^{1}$ and Ludovic Righetti\,$^{1,2}$}
\def\Address{$^{1}$Movement Generation and Control Group, Max Planck Institute for Intelligent Systems, T\"ubingen,  Germany \\
$^{2}$Machines in Motion Laboratory, Tandon  School  of  Engineering,  New  York  University, New  York,  USA  }
\def\corrAuthor{Miroslav Bogdanovic}

\def\corrEmail{mbogdanovic@tue.mpg.de}

\begin{document}
\onecolumn
\firstpage{1}

\title[Generating Robust Policies for Dynamic Locomotion]{Model-free Reinforcement Learning for Robust Locomotion using Demonstrations from Trajectory Optimization} 

\author[\firstAuthorLast ]{\Authors}
\address{}
\correspondance{}

\extraAuth{}

\maketitle

\begin{abstract}

We present a general, two-stage reinforcement learning approach to create robust policies that can be deployed on real robots without any additional training using a single demonstration generated by trajectory optimization.
The demonstration is used in the first stage as a starting point to facilitate initial exploration.
In the second stage, the relevant task reward is optimized directly and a policy robust to environment uncertainties is computed.
We demonstrate and examine in detail the performance and robustness of our approach on highly dynamic hopping and bounding tasks on a quadruped robot.
Accompanying videos can be viewed 
\underline{\href{https://youtu.be/lDPJzpVzLIk}{here}}.
\tiny
 \keyFont{ \section{Keywords:} legged locomotion, deep reinforcement learning, trajectory optimization, robust control policies, contact uncertainty}
\end{abstract}

\section{INTRODUCTION}

\textbf{Trajectory optimization.}
Trajectory optimization (TO) is a powerful tool for generating stable motions for complex and highly constrained systems such as legged robots \citep{winkler2018gait,carpentier2018multicontact,ponton2021efficient}.
However, re-planning trajectories over a receding horizon using full-body dynamics and contacts is still a challenge, because solving a high-dimensional nonlinear program is expensive and can seldom be done in real-time.
Even if it was feasible to re-plan trajectories with sufficient computation, it would be very inefficient to do so in order to reproduce similar motions in similar situations each time.
Furthermore, apart from recent works that explicitly take into account contact uncertainty to design robust control policies \citep{aydinoglu2020stabilization,hammoud2021impedance}, the inclusion of robustness objectives in trajectory optimization can quickly end up in problems that cannot be solved in real time for high-dimensional systems in multi-contact scenarios.

\textbf{Deep reinforcement learning.}
Deep reinforcement learning (DRL) has the potential to resolve some of these issues.
It has recently shown great promises to control complex robotic tasks, e.g. object manipulation \citep{kalashnikov2018qt}, quadrupedal \citep{hwangbo2019learning} and bipedal \citep{xie2020learning} locomotion.
One of the key issues in learning control policies for robotic tasks is exploration.
Discontinuities in the environment arising from contact interactions, in combination with often sparse rewards make it difficult for random exploration approaches to find behaviors necessary for solving a task.

\textbf{Learning from demonstrations.}
An interesting approach to address this issue is to utilize demonstrations for the given task \citep{schaal1997learning,ijspeert2002learning,peters2008reinforcement}.
By providing to the reinforcement learning algorithm basic motions required for completing the task, we remove the need for the algorithm to find it on its own using random exploration.
Here, one can combine trajectory optimization with deep reinforcement learning, by utilizing motions generated by trajectory optimization as demonstrations used to further generate robust policies using deep reinforcement learning.

Demonstrations can be utilized by reinforcement learning approaches in several different ways.
In order to train a reinforcement learning policy to reproduce the demonstration behavior, the majority of approaches have some notion of time in the input of the policy.
Some approaches explicitly give time-indexed states from the demonstration trajectory directly as input \citep{peng2020learning,li2021reinforcement}.
Alternatively, some works train the policy to reproduce the demonstration behavior using only a phase variable in the policy input \citep{xie2020learning,siekmann2021blind}.
Removing any notion of time from the input makes it difficult to train robust policies for real systems \citep{xie2020learning}.

\textbf{Robustness and time-dependence.}
There is however a crucial issue in learning control policies in such a way, in particular in the presence of environmental uncertainties.
As an example, imagine we want to produce a hopping policy that can account for large uncertainties in the ground height.
When contact is made at a different time than in the demonstration, the phase given as input to the policy will be different than the actual phase in the task.
Instead of just going directly into a baseline hopping motion after making contact with the ground, the policy would need to force itself to get back into phase with the demonstration to have any luck to complete the task.
Even removing any notion of time from the input does not on its own solve this issue.
The policy would still need to get back in phase with the demonstration trajectory, but without the time-based inputs lack the information needed to be able to do so.
Hence, fully removing time dependence from the demonstration trajectories in the final feedback policy is key in our approach to provide robustness with respect to contact timing uncertainties.

There are additional benefits in eliminating time dependence when training robust control policies.
It allows us to more broadly randomize initial configurations of the system, something key in deploying learned policies on real robots.
It also separates behavior given in the demonstration from the task goals, allowing us to improve beyond the demonstration for all the important aspects of the task at hand.

\textbf{Our approach.}
In this work we propose a general approach that combines TO and DRL in order to produce robust policies that can be deployed on real robots.
We benefit from trajectories produced by TO to bootstrap DRL algorithms and avoid exploration issues.
We then use DRL to, based on these demonstrations, produce policies robust to environmental uncertainties.
In this way we get the best of both worlds.
Starting from TO trajectories affords solving complex tasks DRL would otherwise struggle with.
The two-stage DRL approach we propose then allows us to avoid the above issues that arise when learning based on demonstrations and learn entirely time-independent robust policies.

\subsection{Related work}

Several recent works have used reinforcement learning to compute policies for locomotion tasks in simulation before deploying them successfully on a real quadruped \citep{hwangbo2019learning, peng2020learning, lee2020learning} or biped robot \citep{xie2020learning, li2021reinforcement, siekmann2021blind}.

In \citep{xie2020learning}, similarly to our work, the authors start by learning a policy to track a reference motion and then further improve it to enable its transfer to the real system.
Crucially, unlike our approach, some notion of time remains present in the policy input in all stages of training, either as a reference motion or a phase variable.
The reward for tracking the original trajectory also remains present in further stages of training, preventing the policy to freely adapt away from it in order to optimize task performance.
Finally, while the resulting policies show some robustness to external perturbations, there are no environment uncertainties present and no need to adapt the timing of the behavior to account for it.

In \citep{li2021reinforcement}, the authors aim to improve upon some aspects of \citep{xie2020learning}.
Instead of learning the policy output in the residual space, i.e. learning only the correction with respect to the demonstration trajectory, they learn the full control signal.
While this makes it easier for the policy to adapt its actions away from the demonstration, the time-dependence is still present in the policy input and the issue of adaptation of timing of the demonstration remains.

In \citep{peng2020learning} the control policy is computed in a single stage of training, by learning to track the given demonstration behavior.
For successful transfer to real robot they rely on domain adaptation, finding the latent encoding over the set of dynamics parameters that performs the best.
There is however no randomization of the environment during training and all test are performed on flat ground.
Similarly to the previous works, the notion of time remains present in the policies here as well, explicitly as a goal given to the policy containing robot states from the reference motion in several of the following time steps.

Unlike these approaches, the most successful recent learning based approach for locomotion in a challenging uncertain environment does not utilize demonstrations at all.
The authors of \citep{lee2020learning} learn a robust locomotion policy for a quadruped robot that performs well on uneven and uncertain surfaces.
The lack of any notion of pre-determined timing affords more room to the policy to adapt to environmental uncertainties (in this case even explicitly by outputting the frequency of the motion).
However, while learning the initial behavior from scratch is possible for the specific task studied in this paper due to the structure imposed by the proposed controller, in many cases it is still necessary to utilize demonstrations to do so.
Therefore, the issue we address in this work, i.e. finding ways to build fully adaptive policies starting from demonstration behaviors, remains a challenge.

\subsection{Contributions}
The main contributions of this paper are as follows:
\begin{itemize}
    \item We propose a framework to exploit the benefits of both TO and DRL to generate control policies that are robust to environmental uncertainties. A key aspect of our framework is to lose time-dependence from the initial trajectories and to build a policy that can adapt to large uncertainties in the environment. 
    \item We implement two highly dynamic motions on a quadruped and show that our framework can deal with random uneven terrains as well as external disturbances. To the best of our knowledge, these results are the first demonstrating the successful use of DRL to robustly realize such behaviors.
\end{itemize}

\begin{figure*}
    \includegraphics[width=\linewidth]{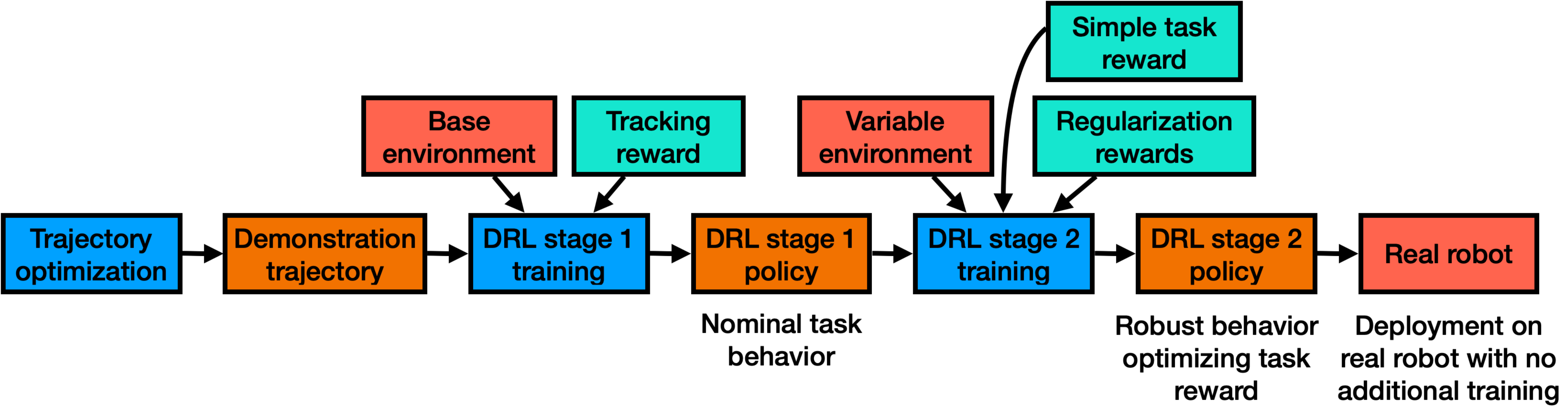}
    \caption{Schematic of our proposed framework. We start from a single demonstration trajectory generated by TO. In the first DRL stage we learn a policy that tracks that demonstration trajectory and successfully produces nominal behavior in simulation. To enable transfer to the real robot, the second DRL stage starts with the resulting policy from the first stage and tries to robustify the policy by randomizing contact and to optimize for performance by replacing demonstration tracking reward with task reward. We directly apply the output policy from the second stage to the robot without any domain randomization of robot parameters.}
    \label{method}
\end{figure*}

\section{PROPOSED ALGORITHM}

\textbf{Algorithm overview.}
Our proposed algorithm has three main parts as shown in Fig.~\ref{method}.
First we use TO to generate efficiently, based on a nominal model of the robot, initial trajectories for new tasks.
The first stage of DRL training then proceeds to build a control policy around this trajectory, caching the solution to a light neural network.
In the second stage of DRL training we replace the trajectory tracking optimization with one that directly optimizes task performance and introduce uncertainties in the training environment.
This allows us to further adapt the policy from the first stage, making it robust and independent from the initial demonstration.
As a final result, we get a policy that can be directly deployed on the real system without any additional training.

\subsection{Trajectory optimization}

\textbf{Generating initial demonstration.}
We use trajectory optimization to generate a nominal motion for the desired task based on a nominal model of the robot and the environment.
In this work, we use the trajectory optimization algorithm proposed in \citep{ponton2021efficient} to compute such demonstrations.
It is important to emphasize that any other trajectory optimization algorithm could also be used, as long as it provides a set of full-body trajectories. 
However, the more realistic the generated motion is, the easier it is for the first stage of DRL to find a policy that tracks it. %
In addition, we do not utilize control actions provided by the demonstration trajectory. Only the state trajectories are required.
This allows us to use any control parametrization for the policy being learned, potentially different than the one used to generate the demonstration. For instance, it is well known that having the policy output the next desired state for a fixed PD controller (rather than torque) is beneficial in terms of transfer to the real world \citep{hwangbo2019learning,bogdanovic2020learning,siekmann2021blind}. However, finding these actions in a trajectory optimization setting is not necessarily trivial.

\subsection{DRL stage 1: Learning a policy to track a given trajectory}

\textbf{Control policy.}
We use a neural network to parametrize the control policy.
A relatively small network proved sufficient for the tasks we considered, with two layers of 64 units each.
We keep the observation space of the policy simple, with the positions and velocities for each joint ($\mathbf{q}^{joint}$, $\dot{\mathbf{q}}^{joint}$) and only robot base variables relevant for the current task.
For the hopping task, this consists of the base position and velocity along the Z-axis ($z^{base}$, $\dot{z}^{base}$).
For the bounding task we also add the angular position and velocity around Y-axis ($
\theta_y^{base}$, $\dot{\theta}_y^{base}$).

The policy outputs parameters for a PD controller in joint space.
It gives desired joint positions at each step, while utilizing fixed P and D gains for control.

Throughout all the stages of training, we use an additional cost term incentivizing the policy to output values for desired joint positions that are actually tracked as well as possible.
Specifically, we penalize the difference between the value given for the desired position by the policy at step $t$ and the actual position achieved at the next step $t + 1$:
\begin{align}
\label{eqn:tt_reward}
\begin{split}
    r_{tt} = -k_{tt} \left\Vert \mathbf{q}^{joint}_{des}(t) - \mathbf{q}^{joint}(t + 1) \right\Vert^2
\end{split}
\end{align}
We note that while this reward term incentivizes a policy to produce a trajectory that is well tracked, it does not prevent it to give values off the trajectory to create forces during contact when necessary.
This has been a crucial aspect of our previous work \citep{bogdanovic2020learning} to enable direct policy transfer to real robots without domain randomization of robot parameters.

\textbf{Training procedure.}
We use Proximal Policy Optimization (PPO) \citep{schulman2017proximal} to optimize the policies, but we do not have many requirements in the choice of the algorithm.
As we use the provided demonstration to resolve exploration issues, we do not need a, potentially off-policy, reinforcement learning algorithm with strong characteristics in this regard.
We instead choose an on-policy algorithm with good convergence properties.

In this stage, the optimized reward consists of two parts: a part for tracking the time-based demonstration and the above-defined regularization term that is a part of the controller parametrization:
\begin{alignat}{2}
\label{eqn:stage1}
\begin{split}
    r_{s1} & =  r_{ti} + r_{tt} \\
    r_{ti} & = k_{ti1} \exp \Big(-{k_{ti2} \lVert \mathbf{x}^{base}_{demo} - \mathbf{x}^{base} \rVert} \Big) \\
               & \quad + k_{ti3} \exp \Big(-{k_{ti4} \lVert \dot{\mathbf{x}}^{base}_{demo} - \dot{\mathbf{x}}^{base} \rVert} \Big) \\
               & \quad + k_{ti5} \exp \Big(-{k_{ti6} \lVert \mathbf{q}^{base}_{demo} \ominus \mathbf{q}^{base} \rVert} \Big) \\
               & \quad + k_{ti7} \exp \Big(-{k_{ti8} \lVert \boldsymbol{\omega}^{base}_{demo} - \boldsymbol{\omega}^{base} \rVert} \Big) \\
               & \quad + k_{ti9} \exp \Big(-{k_{ti10} \lVert \mathbf{q}^{joint}_{demo} - \mathbf{q}^{joint} \rVert} \Big) \\
               & \quad + k_{ti11} \exp \Big(-{k_{ti12} \lVert \dot{\mathbf{q}}^{joint}_{demo} - \dot{\mathbf{q}}^{joint} \rVert} \Big) \\
    r_{tt} & \text{ as defined in \eqref{eqn:tt_reward},}
\end{split}
\end{alignat}
where $\mathbf{x}^{base}$ is the position of the robot base, $\mathbf{q}^{base}$ the base quaternion, $\boldsymbol{\omega}^{base}$ the base angular velocity and $\mathbf{q}^{joint}$ the joint positions. $k_{ti1}$, ..., $k_{ti12}$ represent individual weight and scale constants for each term. We mark difference between two quaternions with $\ominus$.

Following \citep{peng2018deepmimic}, we make two further design choices that prove to be vital in making the training robustly work:
\begin{itemize}
    \item
    We initialize each episode at a randomly chosen point on the demonstration trajectory.
    \item
    We terminate episodes early if the robot enters states that are not likely to be recoverable (based for example on tilt angle of the robot base) or just not conducive for learning (for example knees of the robot making contact with the ground).
\end{itemize}

\textbf{Output.}
In the first DRL stage, we aim to produce a policy that provides some nominal behavior on the task in simulation. However, in our experiments, these policies failed to transfer to the real robot. They remain static, cause shaky behavior on the robot, or result in motions with severe impacts (examples in the accompanying video). To solve these problems, we need an additional stage of training that generates policies that can transfer to the real robot.

\subsection{DRL stage 2: Generating robust time-independent policy}

To create policies that are successfully transferable to the real robot, we continue training starting from the policies outputted from the first stage, introducing the following changes in the training procedure:

    \textbf{Initialization.} We replace initialization on the demonstrated trajectories with initialization in a wider range of states.
    This allows us to better cover the range of states the policy might observe when deployed on the real system, allowing it to learn how to recover and continue the motion in those cases.
    
    \textbf{Environment uncertainties.} We introduce uncertainties in the environment in order to produce more robust motions when deployed on the real system.
    In this work, we are mainly concerned with randomization of the contact surface heights and friction.
    
    \textbf{Time-independent task reward.} We replace the time-based demonstration tracking reward with a time-independent, direct task reward.
    We have already noted the issues that can arise while trying to account for environment uncertainties while tracking a time-based demonstration trajectory.
    The policy is locked into trying to follow the specific time schedule regardless of the environment, whereas adapting it would produce much better recovery behavior.
    Switching to a time-independent reward, directly defining the task helps us deal with this.
    
    Switching to this task reward has additional benefits.
    It allows us to directly optimize desirable aspects of the task, whereas the demonstration only needs to give us some nominal behavior on the task.
    The policy is free to change the behavior in a way that is needed to perform the task in the best possible way, without being penalized for not doing it in the same way as in the demonstration.
    We can also, as we will see later, produce varied behavior starting from a single demonstration by adapting this task reward.

\textbf{Regularization rewards.} Finally, we add additional reward terms in this stage to further regularize the behavior of the learned policies.
We aim to incentivize desirable aspects of policies in the tasks, like torque smoothness and smooth contact transitions.

\section{EVALUATION}

\subsection{Tasks setup}

We evaluate our approach on two different dynamic tasks on a quadruped robot: hopping and bounding.
In both cases we start from a basic demonstration trajectory that provides state trajectories for the current task.
Starting from that we apply our two-step training procedure in simulation to produce robust policies that we then test on a real robot.

\begin{figure*}

      \includegraphics[width=\linewidth]{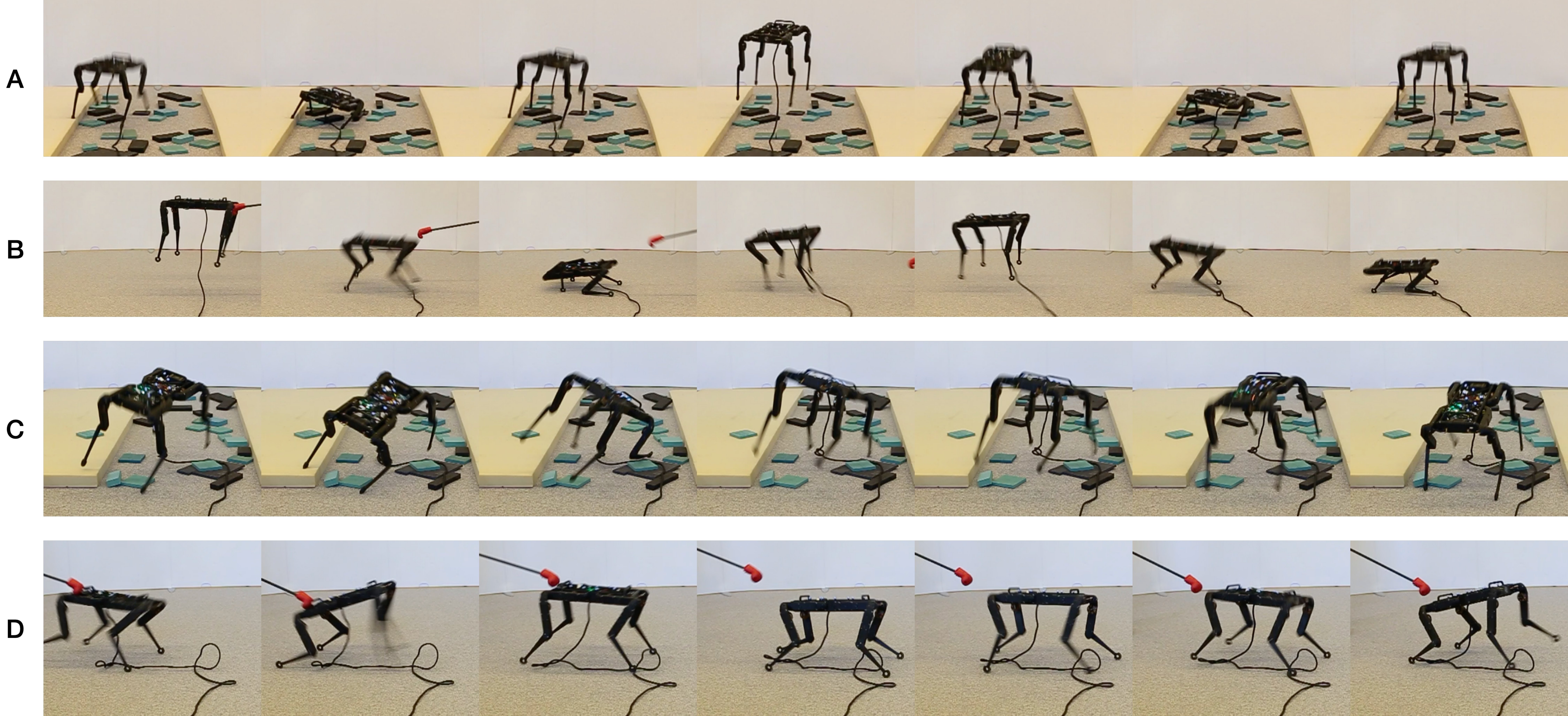}

\caption{Examples of the robustness tests carried out on the quadruped robot Solo8. \textbf{(A)} Hopping on a surface comprised of a soft mattress and small blocks. \textbf{(B)} Hopping with push recovery. \textbf{(C)} Bounding on a surface comprised of a soft mattress and small blocks. \textbf{(D)} Bounding with push recovery.}
\label{fig:experiments}
\end{figure*}

We perform experiments on the open-source torque-controlled quadruped robot, Solo8 \citep{grimminger2020open} (see Fig. \ref{fig:experiments}), which is capable of very dynamic behaviors.
For simulating the system we use PyBullet \citep{coumans2020}.

We use exactly the same training procedure in both cases, with the only differences arising from the need to allow for base rotation around one axis in the bounding task.
This is a particular benefit of the approach we present here -- for a new task we only need a single new demonstration trajectory and a single simple reward term defining the task.

\textbf{DRL stage 1: Early termination.}
The only aspect in the first stage of training specific to the chosen tasks is how we perform early termination.
We perform early stopping here based on the current tilt of the robot base (we increase the range appropriately for the bounding task), as well as when any part of the robot that is not the foot touches the ground.

\textbf{DRL stage 2: Initialization states.}
As noted in the method description, in the second stage of DRL training we introduce a wider range of initialization states.
In the two tasks investigated here, this consists of randomizing the initial height of the base of the robot, tilt of the base around \textit{x}- and \textit{y}-axis and randomness in the initial joint configuration.
We preserve the early termination criteria from stage 1, only extending the range of allowed base tilt angles with the way it is increased in the initialization.

\textbf{DRL stage 2: Environment uncertainties.}
We also introduce uncertainties in the training environment. We randomize the ground position up and down in the range of [\SI{-5}{\cm}, \SI{5}{\cm}] (approximately 20\% of the robot leg length).
We also randomize the ground surface friction coefficient in the range $[0.5, 1.0]$.
While we restrict ourselves only to this limited set of initial state and environment randomizations, as we will see in the later evaluations, this produces policies that are quite robust as they can also handle uneven ground or external perturbations.

\textbf{DRL stage 2: Reward structure (hopping).}
By using a demonstration trajectory to deal with exploration issues, we can define the individual task rewards to be very simple, without the need for any reward shaping.

For the hopping task we use the following reward
\begin{align}
\label{eqn:stage2hopping}
\begin{split}
    r_{s2hp} &= r_{hp} + r_{ps} + r_{ct} + r_{ts} + r_{tt}\\
\end{split}
\end{align}

We use the $r_{hp}$ reward term to define the task
\begin{align}
\label{eqn:stage2hopping}
\begin{split}
    r_{hp} &= \begin{cases}
                  k_{hp} z^{base}, & \text{if } z^{base}_{min} < z^{base} < z^{base}_{max}, \\
                  0, & \text{otherwise}.
              \end{cases} \\
\end{split}
\end{align}
The reward at each timestep is proportional to the current height of the robot base ($z^{base}$), with constant weight $k_{hp}$.
It is clipped to zero below a certain threshold ($z^{base}_{min}$), one that the robot can reach without leaving the ground.
We additionally clip the value of this reward to be zero above a certain height threshold ($z^{base}_{max}$) to incentivize lower hops.
We will additionally vary this threshold to produce policies with different hopping heights starting from the same demonstration trajectory.

We also introduce several reward terms to incentivize different desirable aspects in the resulting behavior.
They reward the base to be close to its horizontal default posture ($r_{ps}$) and smooth contact transitions ($r_{ct}$) and torque smoothness ($r_{ts}$).

We reward the policy for being static in all the base dimensions (positions ($x^{base}$, $y^{base}$) and Euler angles ($\theta_x^{base}$, $\theta_y^{base}$, $\theta_z^{base}$)) except the one the motion is performed on (\textit{z}-axis in this case). With $k_{ps1}$, ..., $k_{ps10}$ being weight and scale constants.
\begin{align}
\label{eqn:stage2hopping}
\begin{split}
    r_{ps} &= k_{ps1} \exp \Big( {-k_{ps2} \lvert x^{base} \rvert ^ 2} \Big) \\
               & \quad + k_{ps3} \exp \Big( {-k_{ps4} \lvert y^{base} \rvert ^ 2} \Big) \\
               & \quad + k_{ps5} \exp \Big( {-k_{ps6} \lvert \theta_x^{base} \rvert ^ 2} \Big) \\
               & \quad + k_{ps7} \exp \Big( {-k_{ps8} \lvert \theta_y^{base} \rvert ^ 2} \Big) \\
               & \quad + k_{ps9} \exp \Big( {-k_{ps10} \lvert \theta_z^{base} \rvert ^ 2} \Big) \\
\end{split}
\end{align}
This term is crucial as it drives the policy to stay at the default posture as much as possible.
Without it the policy could perform the task well in the simulation while always being close to falling over -- which would likely happen when it was transferred to the real system.

The second key reward term asks for smooth contact transition ($r_{ct}$)
\begin{align}
\label{eqn:stage2hopping}
\begin{split}
    r_{ct} &= \begin{cases}
                  -k_{ct} \sum\limits_{i = 1}^4 F_i^{foot}, & \text{if } \sum\limits_{i = 1}^4 F_i^{foot} > F_{limit}^{foot}, \\
                  0, & \text{otherwise}.
              \end{cases} \\
\end{split}
\end{align}
We do so by simply penalizing any contact force values ($F_i^{foot}$) above a certain threshold ($F_{limit}^{foot}$), to penalize impact, with $k_{ct}$ being a constant weight.
Without this term we would have the feet hitting the surface hard on each landing -- this is precisely what we observe in policies from DRL stage 1 where this reward term is not present.
This is not the kind of behavior desired on the real system and these impacts can cause actual damage to the robot.
These types of policies also transfer less well between simulation and real world.
They can learn to perform the task well in simulation by generating hard impacts, but doing so exploits the weaknesses of the contact model in simulation, resulting in a poor performance when transferred to the real system.
Smooth contact transitions enable a better transition between simulation and the real system. Thus, policies which incentivize those better transfer to the real system.

The third reward term ($r_{ts}$) prevents the policy to ask for a very quick change in the desired torque which is not realizable on the real robot with a limited control bandwidth
\begin{align}
\label{eqn:stage2hopping}
\begin{split}
    r_{ts} &= -k_{ts1} \exp \Big( {k_{ts2} \left\Vert \boldsymbol{\tau}(t) - \boldsymbol{\tau}(t - 1) \right\Vert} \Big) \\
\end{split}
\end{align}
with $\tau$ being the joint torque and $k_{ts1}$ and $k_{ts2}$ weight and scale constants respectively.

The final reward term $r_{tt}$ is the same one we use in the first stage of training (as defined in \eqref{eqn:tt_reward}).

\textbf{DRL stage 2: Reward structure (bounding).}
For the bounding task, we only make changes to the parts of the reward defining the task:
\begin{align}
\label{eqn:stage2bounding}
\begin{split}
    r_{s2bn} &= r_{bn} + r_{cc} + r_{ps} + r_{ct} + r_{ts} + r_{tt}\\
\end{split}
\end{align}

We define the task reward here in two parts, $r_{bn}$ and $r_{cc}$.
$r_{bn}$ rewards the policy for being close to the path the demonstration takes in the $[z^{base}, \theta_y^{base}]$ space
\begin{align}
\begin{split}
    r_{bn} &= -k_{bn} \min_i \lVert (z^i - z^{base}, \theta_y^i - \theta_y^{base}) \rVert \\
\end{split}
\end{align}
with $k_{bn}$ being a constant weight.
We do this with no concept of time in this case, by just taking the distance to the closest point.
This gives the policy freedom to perform the motion slower or faster, with different amplitude.
We will later see that this results in a variety in bounding behaviors from repeated trainings, independent of the timing in the original demonstration.

The second part of the task reward, $r_{cc}$, is related to the contact state
\begin{align}
\begin{split}
    r_{cc} &= \begin{cases}
                  k_{cc}, & \text{only front two legs in contact}, \\
                  k_{cc}, & \text{only back two legs in contact}, \\
                  k_{cc}, & \text{no legs in contact}, \\
                  0, & \text{otherwise}.
              \end{cases} \\
\end{split}
\end{align}
with $k_{cc}$ being a constant reward.
It incentivizes the policy to, when making contact with the ground, only do so with front or back legs at the same time.
Without anything to incentivize the policy to do this, we have observed DRL stage 1 policies reproducing the bounding motion while keeping all feet in contact with the ground.
This reward part ensures appropriate contact states with flight phases in between.

We keep the other reward terms, ones used to incentivize desired aspects of the behavior, the same as in the hopping task ($r_{tt}$ as defined in \eqref{eqn:tt_reward}, $r_{ct}$, $r_{ts}$ as defined in \eqref{eqn:stage2hopping}).
The one change we make is to the reward incentivizing the robot base staying close to the default posture, $r_{ps}$
\begin{align}
\label{eqn:stage2bounding}
\begin{split}
    r_{ps} &= k_{ps1} \exp \Big( {-k_{ps2} \lvert x^{base} \rvert ^ 2} \Big) \\
               & \quad + k_{ps3} \exp \Big( {-k_{ps4} \lvert y^{base} \rvert ^ 2} \Big) \\
               & \quad + k_{ps5} \exp \Big( {-k_{ps6} \lvert \theta_x^{base} \rvert ^ 2} \Big) \\
               & \quad + k_{ps7} \exp \Big( {-k_{ps8} \lvert \theta_z^{base} \rvert ^ 2} \Big) \\
\end{split}
\end{align}
We do not reward staying at default posture in the $\theta_y^{base}$ direction in this case, as that is the angle the robot is moving around while bounding.

\subsection{Hopping task results}

\begin{figure*}
    \includegraphics[width=\linewidth]{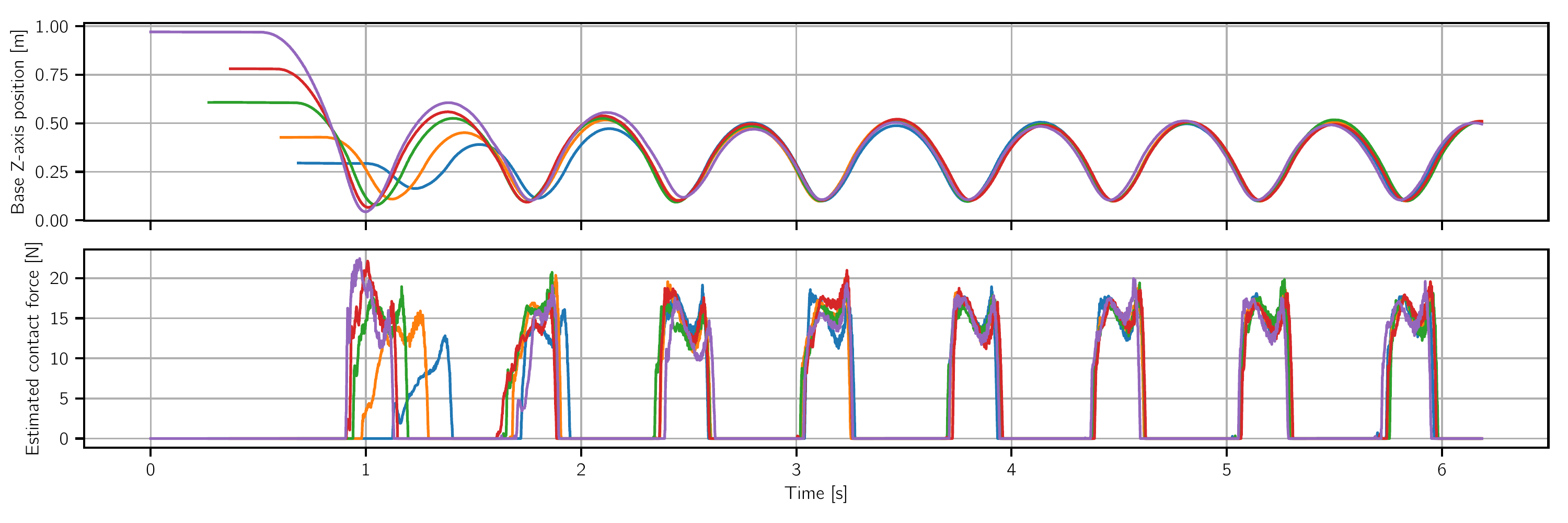}
    \caption{Hopping experiments started from different initial heights. The top figure shows that after dropping the robot from a set of different heights ranging between 0.3-1 m, the robot quickly goes back to the nominal behavior which is hopping with the maximum height of 0.5 m. The bottom figure shows that, dropping from different initial heights, the robot is able to adapt its landing such that the impact forces remain very low and almost invisible in the estimated force. }
    \label{hopping_varible_height}
\end{figure*}

Figure \ref{hopping_varible_height} shows results when the real robot is dropped from different heights to start the motion.
We show the base height and estimated contact force for one of the legs as a function of time.
As we do not have force sensors in the feet, we estimate the contact forces based on the torques the robot applies, using $F_i^{foot} = (S_i J_i^{T})^{-1} \, S_i \, \tau$, where $S_i$ and $J_i$ are the joint selection matrix of the leg $i$ and Jacobian of the foot $i$, respectively. Note that this estimation ignores the energy dissipated through damping of the robot structure and drive system. However, it provides an approximate measure of contact forces sufficient for the analyses of this paper.
We further align the plots based on the later part of the motion -- the stable cycle the robot gets into.

First, we can note that regardless of the drop height the robot goes into the same stable hopping cycle.
What is more, it does so very quickly, as we can see all the individual rollouts matching after only two hops.
We can also note here the benefits of time independence of the policy.
It is what allows us to be able to start the motion from this large range of initial heights.
It is also what enables this fast stabilization, as we can see that the two initial hops are on a different cycle -- one needed to stabilize the motion properly.

This test also highlights the general quality of contact interaction achieved with this approach.
We can see that the impact forces, even on the highest drop (\SI{1}{\m} height), barely go over the force values for the stable hopping cycle (\SI{50}{\cm} height).
This is purely learned behavior, as a result of impact penalties introduced in stage 2 of DRL training.
It is not present in the demonstration and when we test DRL stage 1 policies on the real system high impact forces are generated and the policies are very fragile.
Smooth contact transitions can also be observed in the accompanying video.

Further, the behavior is robust to uneven terrain and external pushes although this was never explicitly trained for.
The robot is able to recover from significant tilt of the base arising from either external pushes or landing on an uneven surface (Fig.~\ref{fig:experiments}A, \ref{fig:experiments}B).
More extensive examples of recovery behavior can be seen in the accompanying video.

Finally, we demonstrate the variety of robust behaviors that can be optimized from the same demonstration by doing repeated trainings with different values of the $z^{base}_{max}$ threshold in the hopping task reward.
With this simple change in the task reward, starting from one demonstration, we can produce hopping behaviors at different heights.
Examples of this can be seen in the accompanying video.

\subsection{Bounding task results}

\begin{figure}
  \includegraphics[width=\linewidth]{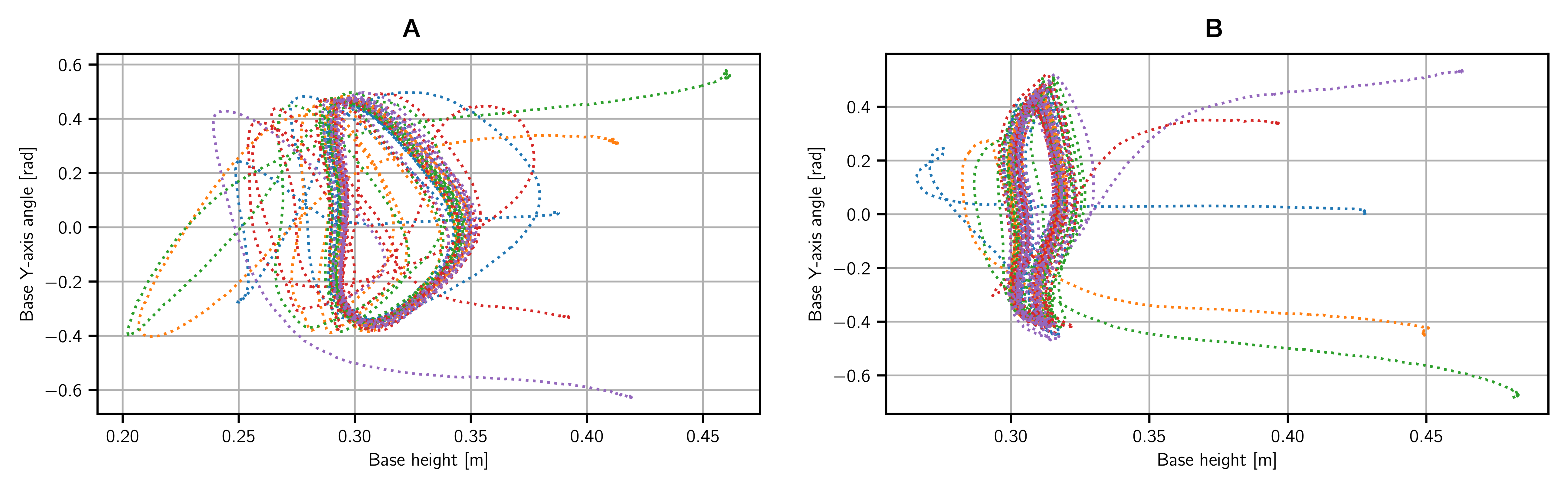}

\caption{Two different bounding behaviors on the robot with different initial conditions. Starting with a wide range of initial angles for the base in y-direction (roughly between -35 to 35 deg), the robot quickly converges back to the desired behavior.}
\label{fig:bounding_variable_drop_angle}
\end{figure}

In Fig.~\ref{fig:bounding_variable_drop_angle} we show results for a test where we drop the robot from different angles to start the motion.
We perform the same test for two different final DRL stage 2 policies for this task.
We can see that the policies can handle a wide range of initial base angles -- around 35 degrees in both directions.
What is more, as was the case with the hopping task, we can see that here as well all the initializations end up in the same stable motion cycle.

The bounding motion exhibits similar robustness to uneven terrain and external perturbations as the hopping motion (Fig.~\ref{fig:experiments}C, \ref{fig:experiments}D).
Same as with the hopping task, the policies rely on their knowledge of how to handle a varied set of base states to recover from anything that arises from these conditions even through it was not explicitly trained for.

In this task, we would also like to demonstrate the variety of behaviors we can generate from one single demonstration.
Unlike in the hopping task, where we made simple changes in the task reward to achieve different jumping heights, we instead give more freedom to the task reward and examine the variety of produced behaviors.
As noted in the task reward definition, the policy has the freedom to produce slower or faster bounds with smaller or larger amplitudes.
As seen in the accompanying video we arrive at a variety of bounding behaviors in this way, starting from the same initial demonstration trajectory.

\section{DISCUSSION}

\textbf{Simplicity and generality of the approach.}
One of the main benefits of the approach is its generality and simplicity.
The only elements needed for each new task, as can be seen from the two tasks studied here, are a single demonstration trajectory and a direct straightforward task reward.

The demonstration does not need to provide ideal performance on the task, as we optimize task performance in the later stage of training.
It simply needs to provide a sequence of states, and not necessarily actions, that enable a decent performance. Providing only states is also simpler in cases where it is not trivial to calculate the exact forces to realize a particular motion.

As for the task reward, reward shaping is not needed, as the demonstration resolves any exploration issues that could occur as a result of sparse reward signal.
We can directly reward aspects of the task of interest.
We keep reward terms other than the task reward as general as possible, encoding characteristics of general good robot behavior. We expect those to remain constant across a varied range of tasks.

\textbf{Reinforcement learning perspective.}
From the reinforcement learning perspective our approach presents a simple and effective way to deal with exploration issues in robotic tasks.
We also remove the trajectory tracking reward in the second stage of our training, so, as seen in our experiments, we are able to change the policy away from the exact behavior defined in the demonstration.

\textbf{Trajectory optimization perspective.}
From the trajectory optimization perspective, our approach proposes a systematic way to consider different types of uncertainty and find a robust control policy for robotic tasks, especially those with contact. Furthermore, our approach caches the solution of a model-based approach for future use and eliminates the need for re-generating repetitive motions. We believe this is a practical way to combine the strength of trajectory optimization and reinforcement learning for continuous control problems; 1) Trajectory optimization is used to generate a desired behavior efficiently to achieve the task at hand  2) different types of realistic uncertainties are easily added to the simulation, e.g. contact timing uncertainty, and DRL is used to produce a robust feedback policy.

\section{CONCLUSION}

In this work, we presented a general approach for going from trajectories optimized using TO to robust learned policies on a real robot.
We showed how we can start from a single trajectory and arrive at a robust policy that can be directly deployed on a real robot, without any need for additional training.
Through extensive tests on a real quadruped robot, we demonstrated significant robustness in the behaviors produced by our approach.
Importantly, we do so in setups, uneven ground and external pushes, for which the robot was not explicitly trained for.
All this gives hope that such approaches could be used across varied robotic tasks to simply generate robust policies to be used on real hardware, bridging the gap between trajectory optimization and reinforcement learning in such tasks.

\section*{Conflict of Interest Statement}

The authors declare that the research was conducted in the absence of any commercial or financial relationships that could be construed as a potential conflict of interest.

\section*{Author Contributions}

M.B., M.K., and L.R. designed research; M.B. performed numerical simulations; M.K. prepared the demonstrations; M.B. and M.K. performed hardware experiments; M.B., M.K. and L.R. analyzed the results and wrote the paper.

\section*{Funding}
This work was supported by New York University, the European Union’s Horizon 2020 research and innovation program (grant agreement 780684) and the National Science Foundation (grants 1825993, 1932187 and 1925079).

\section*{Acknowledgments}
We would like to thank the contributors of the Open Dynamic Robot Initiative (ODRI) for the development of the hardware, electronics and the low-level software for controlling the robot Solo.

\bibliography{example} 
\bibliographystyle{frontiersinSCNS_ENG_HUMS}

\end{document}